\documentclass[letterpaper, 10 pt, conference]{ieeeconf}  % Comment this line out if you need a4paper

\IEEEoverridecommandlockouts                              % This command is only needed if 
\usepackage{tikz}

\usepackage{graphicx}

\usepackage{amsmath}
\usepackage{amssymb}
\usepackage{booktabs}
\usepackage{xcolor}
\usepackage{mathtools}
\usepackage{float}
\usepackage{xspace}
\usepackage{cite}
\usepackage{caption}
\usepackage{subcaption}
\usepackage[inline]{asymptote}
\usepackage{tikz}
\usepackage{makecell}
\usepackage{comment}
\usepackage{multirow}
\usepackage{multicol}
\usepackage{booktabs}
\usepackage{pifont}
\let\labelindent\relax
\usepackage{pgfplots}
\usepackage{nomencl}
\usepackage[english]{babel}
\usepackage[autostyle]{csquotes}
\usepackage[inline]{enumitem}
\usepackage{amsthm}
\usepackage{stackengine} 
\usepackage{resizegather}

\usepackage[pagebackref,breaklinks,colorlinks]{hyperref}

\usepackage[capitalize]{cleveref}
\usepackage{pgfplots}
\usepackage{tikz}

\crefname{section}{Sec.}{Secs.}
\crefname{table}{Tab.}{Tabs.}
\pgfplotsset{width=10cm,compat=1.9}

\usepgfplotslibrary{external}

\DeclareMathOperator*{\argmin}{arg\,min}

\def\lidar{\texttt{LiDAR}\xspace}
\def\sota{\texttt{SOTA}\xspace}
\def\dem{\texttt{DEM}\xspace}
\def\dems{\texttt{DEMs}\xspace}
\def\dof{\texttt{DOF}\xspace}
\def\ldc{\texttt{LDC}\xspace}
\def\sothree{\texttt{SO(3)}\xspace}
\def\sotwo{\texttt{SO(2)}\xspace}
\def\setwo{\texttt{SE(2)}\xspace}
\def\sethree{\texttt{SE(3)}\xspace}
\def\dyt{\texttt{DYT}\xspace}

\def\slam{\texttt{SLAM}\xspace}
\def\ransac{\texttt{RANSAC}\xspace}
\def\icp{\texttt{ICP}\xspace}
\def\cpc{\texttt{CPC}\xspace}
\def\cnn{\texttt{CNN}\xspace}

\newcommand{\myfirstpara}[1]{\noindent \textbf{#1}:}
\newcommand{\mypara}[1]{\vspace{0.5em} \myfirstpara{#1}}

\newcommand{\cmark}{\textcolor{green}{\checkmark}}%
\newcommand{\xmark}{\textcolor{red}{\ding{55}}}%

\title{\LARGE \bf
FinderNet: A Data Augmentation Free Canonicalization aided Loop Detection and Closure technique for Point clouds in 6-DOF separation.
}

\author{ Sudarshan S Harithas$^{1}$ ,Gurkirat Singh$^{1}$, Aneesh Chavan$^{1}$, Sarthak Sharma$^{1}$, Suraj Patni$^{2}$, \\ Chetan Arora$^{2}$, K. Madhava Krishna$^{1}$  % <-this % stops a space
\thanks{$^{1}$are with RRC, IIIT Hyderabad, India
        }%
\thanks{$^{2} $is with University of IIT Delhi, India }%
\thanks{$^\dagger$Project page: \href{http://gurkiratsingh.me/FinderNet/}{https://gurkiratsingh.me/FinderNet/} } 
}

\begin{document}

\maketitle
\thispagestyle{empty}
\pagestyle{empty}

%%%%%%%%%%%%%%%%%%%%%%%%%%%%%%%%%%%%%%%%%%%%%%%%%%%%%%%%%%%%%%%%%%%%%%%%%%%%%%%%
\begin{abstract}

We focus on the problem of \lidar point cloud based loop detection (or Finding) and closure (\ldc) in a multi-agent setting. State-of-the-art (\sota) techniques directly generate learned embeddings of a given point cloud, require large data transfers, and are not robust to wide variations in 6 Degrees-of-Freedom (\dof) viewpoint. Moreover, absence of strong priors in an unstructured point cloud leads to highly inaccurate \ldc. In this original approach, we propose independent roll and pitch canonicalization of the point clouds  using a common dominant ground plane. Discretization of the canonicalized point cloud along the axis perpendicular to the ground plane leads to an image similar to Digital Elevation Maps (\dems), which exposes strong spatial priors in the scene. Our experiments show that \ldc based on learnt embeddings of such \dems is not only data efficient but also significantly more robust, and generalizable than the current \sota. We report significant performance gain in terms of \textit{Average Precision} for loop detection and absolute translation/rotation error for relative pose estimation (or loop closure) on \textit{Kitti}, \textit{GPR} and \textit{Oxford Robot Car} over multiple \sota \ldc methods. Our encoder technique allows to compress the original point cloud by over $830$ times. To further test the robustness of our technique we create and opensource a custom dataset called \textit{Lidar-UrbanFly Dataset (LUF)} which consists of point clouds obtained from a \lidar mounted on a quadrotor.

% an (Average Precision for loop detection, mean absolute translation/rotation error) improvement of (10.5, 16.7/5.43)\% on the \textit{KITTI08} sequence, and (11.0, 34.0/25.4)\% on \textit{GPR10} sequence, over the current \sota.   

% We create a novel 6-\dof experiment to test the generalizabiliy and view point invariance of our model, here the point clouds are subjected to random rotations between $10-60$ degrees in \textit{roll, pitch and yaw} before \ldc. In the presence of such rotations the \ldc performance of \sota algorithms reduce by more than 90\%, whereas the proposed framework does not incur any such drop.
\begin{figure}[t]
	\includegraphics[width=1.0\linewidth]{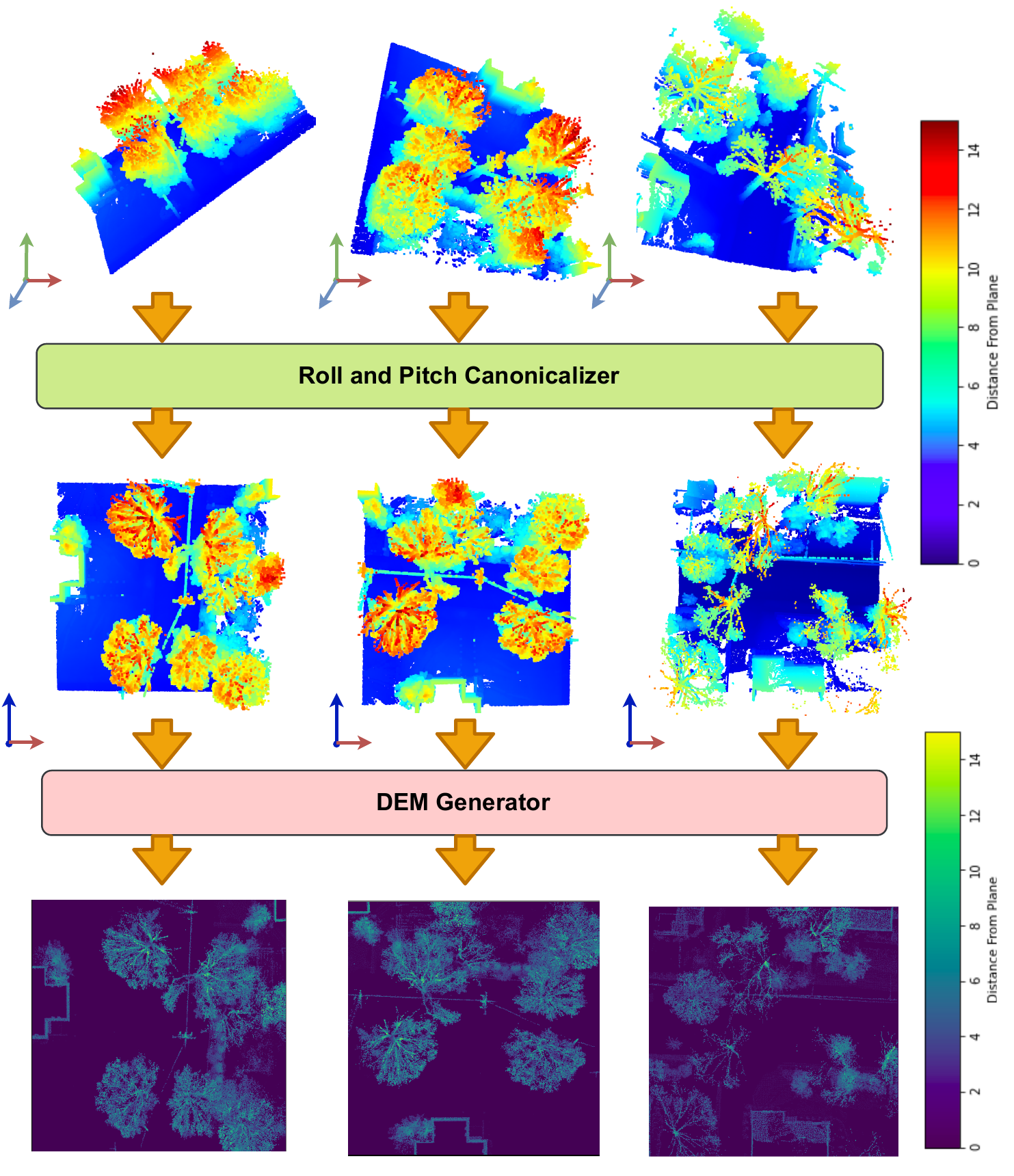}
	\vspace{-1.5em}
	\caption{ \footnotesize{ We observe that raw \lidar point clouds (first row) lack spatial structure for robust loop detection and closure (\ldc). We perform local roll and pitch canonicalization (second row), followed by discretization along the $z$-axis (third row), which leads to output similar to digital elevation maps (\dems) and exposes rich scene structure in the input. Our model performs \ldc on such \dems, leading to high data efficiency, robustness, and generalizability to 6-\dof viewpoint variations.}}
	\label{fig:teaser}
	\vspace{-7mm}
\end{figure}

\end{abstract}

%%%%%%%%%%%%%%%%%%%%%%%%%%%%%%%%%%%%%%%%%%%%%%%%%%%%%%%%%%%%%%%%%%%%%%%%%%%%%%%%
\section{INTRODUCTION}

Loop detection and closure is a critical module in  \slam (\textit{Simultaneous Localization And Mapping}) pipeline to reduce accumulated drift in the estimation process. 
%Recovering the closest possible match between a given query point cloud and a pre-built database is known as \emph{Loop Detection} (or place recognition), whereas the process of estimating the relative transform between the query and the retrieved sample is known as \emph{Loop Closure}. 
As single-agent \slam systems have matured, the research community is increasingly focusing on multi-agent scenarios, and collaborative \slam \cite{door_slam,collab_Slam_survey}. Since, transmitting large point clouds between agents is impractical, judicious use of the limited bandwidth is necessary for distributed Loop Detection and Closure (\ldc) \cite{collab_Slam_survey, scaramuzza_efficient, scaramuzza_efficient2}. This paper focuses on robust, and data efficient distributed \ldc. Though, most components of our pipeline applies to generic point clouds, we assume \lidar as the essential sensing modality.

The techniques are broadly split into two styles: 
\begin{enumerate*}[label=\textbf{(\arabic*)}] 
% \item \textit{Place Recognition (or Loop Detection) only} \cite{pcan, soe_net, pointnetvlad } these methods aim to generate view-invariant point clouds descriptors for global localization in a pre-built map.  
\item \textit{ Loop Detection } Such approaches typically use place recognition methods \cite{pcan, soe_net, pointnetvlad, retriever} to detect the loop and employ traditional point cloud registration algorithms such as \cite{icp, teaser} to estimate the relative pose between the query and the recovered point clouds. 
\item  \textit{ Loop Detection and Closure } Methods such as \cite{LCDNet, overlapnet, oreos} estimate perform place recognition and estimate the relative pose between the query and the retrieved point cloud in an end-to-end pipeline without employing any external point cloud registration methods.
\end{enumerate*} Our approach belongs to second category where we detect and close the loop as a part of a single pipeline without employing any external point cloud registration method. 

Typically, these methods \cite{LCDNet, soe_net, pcan, pointnetvlad,retriever} either depend upon a combination of feature aggregation and data augmentation where they apply randomly sampled rigid transforms to the input point clouds to achieve viewpoint invariance or perform \ldc through overlap estimation. Such a training procedure does not generalize to wide viewpoint variations. We take an original approach where a combination of  canonical representations and differentiable latent space alignment is used to geometrically constrain view invariance into the system. Such a technique leads to \sota performance on multiple datasets.

The cornerstone of our efforts is a \textit{Roll and Pitch (RP) canonicalizer} and a \textit{Differentiable Yaw Transformer} (DYT). The \textit{RP Canonicalizer} makes use of the dominant ground plane hypothesis 
(commonly encountered in autonomous driving and drone applications) 
\cite{ point_to_plane , scan-context, overlapnet,oreos, bv-match} to compensate for the roll and pitch between two point clouds. The roll and pitch canonicalized point clouds are converted into a \textit{Digital Elevation Map} (\dem), a visual explanation of the process is given in Fig. ~\ref{fig:teaser}. We further develop a \textit{Differentiable Yaw Transformer} (\dyt) that
operates on the latent feature embeddings of the \dem to achieve yaw invariance, and provide viewpoint invariant loop detection with 6-\dof (\sothree) relative motion. This is in contrast to existing techniques focusing only on yaw rotation \cite{scan-context,DISCO,overlapnet,scancontext++}.    

\vspace{-2mm}
\mypara{Contributions}  
%The key contributions include:
\begin{enumerate*}[label=\textbf{(\arabic*)}]
\item \textbf{Novel Pipeline:} Instead of directly operating on the raw point clouds that inherently lack structure, we convert the point clouds into a regularly spaced \dem via a roll and pitch canonicalizer (\cref{subsec:dem_generate}) with only the yaw to further deal with.  The canonicalized \dem representation provides a structure that \cnn backbones can readily process, bypassing equivariance issues that typically plague point cloud representations. While \textit{Pointnet} \cite{pointnet} and its variants \cite{pointnet++} handle equivariance, the superiority of the proposed pipeline over \textit{Pointnet} inspired architectures \cite{pointnetvlad, pcan, soe_net} is tabulated in the Results Section over a diverse set of \ldc related performance metrics. 

% \item \textbf{Multi Functionality:} By enabling \ldc on the compressed latent space, we provide for low bandwidth data transfer, which is critical in a multi-agent setting. The querying agent is not required to transmit its entire point cloud, but only its compressed latent embedding. Moreover, the proposed architecture provides for multiple functionalities such as Low Bandwidth Data Efficiency, 6-\dof invariance and \ldc unlike prior art.   
\item \textbf{Canonicalization Pipeline \& Differentiable Alignment for view invariance} Unlike previous methods that approach \ldc through feature aggregation \cite{soe_net, pcan, pointnetvlad,LCDNet} or overlap estimation \cite{overlapnet, overlaptransformer}, we approach \ldc through a canonicalization and differentiable alignment procedure , that enables us to geometrically constrain view invariance into the network and enables training with no data augmentation, and achieves \sota results on multiple datasets. Our latent space alignment is driven by the \dyt which is a novel parameter estimation module which is used for differentiable grid sampling. In contrast to methods such as \cite{overlapnet,oreos} the \dyt allows us to estimates the relative yaw in a self supervised manner, i.e. it does not require explicit supervision of the relative yaw between the two point clouds. 
\item \textbf{6-\dof recovery:} Unlike previous approaches that show loop closure only as a \setwo alignment, the proposed method recovers 6-\dof pose  between the two candidate point clouds, even as it precludes the need for data augmentation, exploiting the inherent viewpoint invariance of the pipeline. 
The proposed framework goes beyond \sota on a number of public datasets such as KITTI \cite{kitti}, GPR \cite{ALITA} and Oxford RobotCar \cite{oxford_robocar} on established performance metrics for \ldc. Specifically, the proposed framework is the best performing on 6-\dof pose recovery and it outperforms most prior art on the \setwo \ldc task.
\end{enumerate*} 
\cref{tab:comparison_rel_work} gives a conceptual comparison of our method with contemporary techniques, where \textbf{DE} refers to methods that perform compression with a downstream objective of performing loop detection or registration in the compressed space, furthermore, these methods are capable of decompressing the point cloud from the compressed space either in its partial or complete form (\dem is a partial reconstruction and a dense result can be obtained by up-sampling and completion). \textbf{LD} are methods that perform the task of loop detection and employ an external point cloud registration method \cite{icp, teaser} to estimate the relative pose. \textbf{LDC} are methods that jointly estimates the loop and the relative pose through a single pipeline. \textbf{NDA} is set to true (or \cmark) when a method can learn without data augmentation and \textbf{VI} are methods that can handle \sethree motion. 
\vspace{-2mm}

\begin{table}[t]
\scriptsize
\vspace{3mm}
	\centering
%	\resizebox{\linewidth}{!}{
	\setlength{\tabcolsep}{8pt}
		\begin{tabular}{llccccc}
			\toprule
			\textbf{Method} & \textbf{Venue} & \textbf{DE} & \textbf{VI} & \textbf{NDA} & \textbf{LD} & \textbf{LDC} \\
			\midrule
			\cite{overlapnet} & RSS'20  & \xmark & \xmark & \xmark & \cmark & \cmark(Yaw)  \\
			\cite{scan-context} & IROS'18& \xmark & \xmark & NA & \cmark & \cmark(Yaw) \\
			\cite{dcpcr} & RAL'22 & \cmark & \cmark & NA & \xmark & \xmark \\
			\cite{retriever} & ICRA'22 & \cmark & \cmark & \xmark & \cmark & \xmark \\
			\cite{deep-compression} & RAL'21 & \cmark & \cmark & \xmark & \xmark & \xmark\\
			\cite{LCDNet} & TRO'22 & \xmark & \cmark & \xmark & \cmark & \cmark \\
			\cite{oreos} & IROS'19 & \xmark & \xmark & \xmark & \cmark & \cmark(Yaw) \\
			\cite{pointnetvlad} & CVPR'18& \xmark & \cmark & \xmark & \cmark & \xmark\\
			\cite{pcan} & CVPR'19 & \xmark & \cmark & \xmark & \cmark & \xmark \\
			\cite{soe_net} & CVPR'21 & \xmark & \cmark & \xmark & \cmark & \xmark \\ 
			\cite{dh3d} & ECCV'20 & \xmark & \cmark & \xmark & \cmark & \cmark \\
			\midrule
			Ours & ****'23 & \cmark & \cmark & \cmark & \cmark & \cmark  \\
			\bottomrule
		\end{tabular} 	
%	}
	\caption{Acronyms: \textbf{DE:} Data Efficiency though learnt embeddings, \textbf{VI:} 6-\dof View Invariance, \textbf{NDA:} No large Data Augmentation requirement, \textbf{LD:} Loop Detection, \textbf{LDC:} Joint Loop Detection and Closure, \textbf{NA:} Not Applicable.}  
	\label{tab:comparison_rel_work}
 \vspace{-6mm}
\end{table}

\begin{figure*}[t]
	\includegraphics[width=1.0\linewidth]{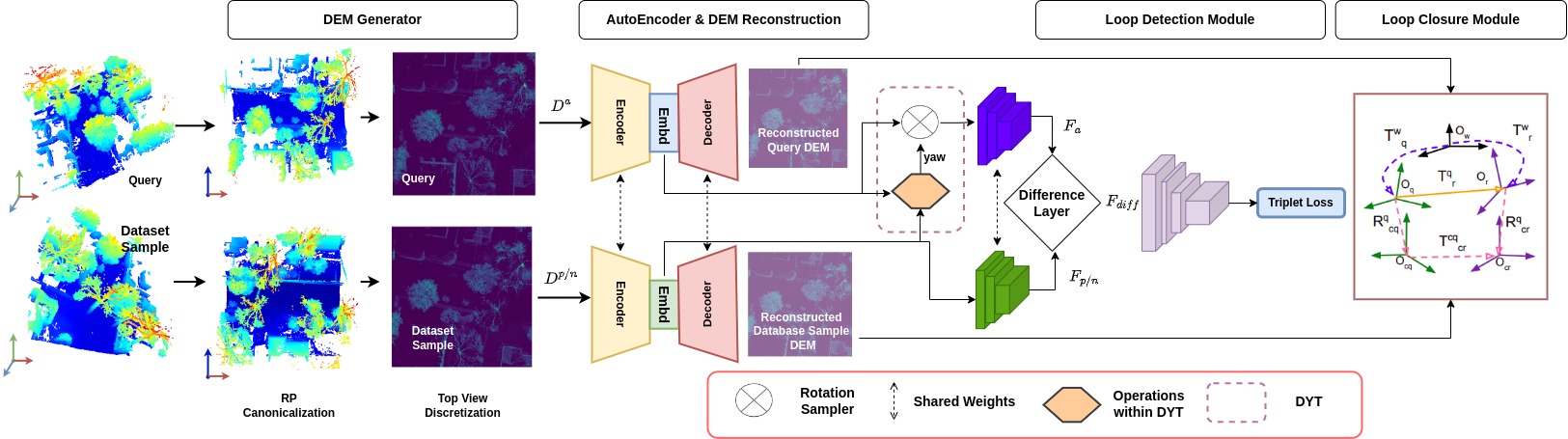}
	\caption{\footnotesize The figure demonstrates the overview of our pipeline; the two point clouds in the extreme left are the input query and database sample; the \textit{DEM Generator} (section \ref{subsec:dem_generate}) generates a discredited top view of the point cloud; and the autoencoder structure further compresses the \textit{DEM} (section \ref{subsec:dem_encoder_decoder}). The \textit{Differentiable Yaw Transformer} (DYT) (section \ref{subsec:dyc}) is used for the yaw alignment, the operations within the \dyt include \textit{CPC}, \textit{Horizontal padding of polar embedding}, and \textit{Correlation}; each of these are explained in section \ref{subsec:dyc}. The complete set of operations is shown as a single orange hexagon; the result of these operations is a scalar yaw value, which is fed into the rotation sampler. We design a network to perform loop detection (section \ref{subsec:loop_detection}) and closure (section \ref{subsec:loop_closure}) using these compressed embeddings without the need for explicit decompression.    }
	\label{overall-flow}
\end{figure*}

\section{Related works}

\myfirstpara{Handcrafted Feature Descriptors} 
%
%Approaches like 
\cite{scan-context, bv-match,lidar-iris} rely on handcrafted feature descriptors to extract local geometric information and aggregate it to obtain a global descriptor suitable for loop detection and closure. \cite{scan-context, bv-match} assume presence of a dominant ground plane to detect and close loops. \cite{scan-context} follows a polar representation, where the ground is discretized into bins by splitting in both radial and azimuthal directions, and each bin storing the maximum height present in the vertical volume. \cite{bv-match} discretizes the ground plane into rectangular cells in a Cartesian form, and each cell storing it's point cloud density. Our representation of \dem is a discrete Cartesian representation of the ground plane, where each grid cell stores the maximum height of the points present in it. However, unlike \cite{bv-match,scan-context}, we perform a complete 6-\dof estimation and loop closure.

% However, they differ in their approach towards representing the discrete ground plane.
%and storing the density i.e the number of points present in the vertical column in each cell. 
% Our approach is similar to \cite{bv-match} as the \textit{DEM} is a discrete cartesian representation of the ground plane, and as \cite{scan-context} we store the maximum height of the points present in the vertical volume of the cell.  It is further important to note that both \cite{bv-match} and \cite{scan-context} do-not perform a 6DOF loop closure, whereas through our \textit{DEM} generation procedure a 6DOF relative pose can be estimated.       

%\textbf{Learning Based Approaches}: Post the breakthroughs in learning based methods fo 2D vision tasks, the research community has been actively developing architectures that deal directly with point clouds for loop detection and closure.
% more and more researcher has focused on representing 3D data such as point clouds with deep learning. 
% The methods either solely perform loop detection or jointly perform loop detection and closure. We review a few methods for either case below. 
% \vspace{-18pt}
\vspace{-1mm}

\mypara{Learning Based Approaches for Loop Detection}
PointNet \cite{pointnet} proposes a  neural network model that directly consumes point clouds while maintaining permutation invariance. \textit{PointNetVLAD} \cite{pointnetvlad} uses \cite{pointnet} and NetVLAD\cite{netvlad} to generate global descriptors for place recognition. \textit{PCAN} \cite{pcan} uses \cite{pointnet} as the backbone architecture to extract local features and the corresponding attention maps along with \cite{netvlad} for feature aggregation. However, both \cite{pointnetvlad,pcan} uses PointNet as a backbone architecture, which processes each point separately via a MLP, not capturing local neighbourhood information. Recently, Retriever\cite{retriever} detects loops directly in the compressed feature representation using Perceiver \cite{perceiver} based mechanism to aggregate the local features. All the above methods use aggregated local features, to compute a global descriptor that is viewpoint invariant, such methods require expensive data augmentation. We propose a canonicalization procedure in order to explicitly enforce viewpoint invariance and in contrast to \cite{pointnetvlad, pcan, retriever} which only perform loop detection, our method performs \ldc.

    % \cite{netvlad} has shown significant progress for 2D image-based loop detection. 
    % The emergence of attention-based learning  \cite{attention} allowed for the development of a wide range of architectures. 
    %which essentially does not concern with the local geometry. 
    %depend upon 3D local feature extraction and aggregation to generate the global descriptor. 
    %However, \cite{retriever} only performs loop detection.  
    % \vspace{-24pt}
\vspace{-7pt}
\mypara{Learning Based Approaches for \ldc}
LCDNet \cite{LCDNet} proposes an end-to-end trainable system, with a \emph{Unbalanced Optimal Transport} algorithm to estimate 6-\dof relative transform between two point clouds. DH3D \cite{dh3d} aggregates local features using hierarchical network to obtain global features for loop detection. %The local features are further used for 6-\dof pose estimation. 
Both \cite{LCDNet, dh3d} rely on  an expensive 6-\dof data augmentation of the input point cloud in order to achieve orientation invariance. %Our proposed roll-pitch canonicalization and yaw transformer enables to forego the expensive data augmentation step by aligning the roll-pitch-yaw in the input point clouds, making all the successive operations orientation invariant. 
The proposed framework bypasses data augmentation through explicit roll-pich canonicalization followed by yaw alignment.
%It also enables us to preserve the original point cloud, and ultimately estimating the relative 6DOF. 
Unlike \cite{LCDNet, dh3d}, we operate on highly compressed point cloud representation, making our approach suitable for data transmission in a multi-agent setting. Overlap-based approaches such as OverlapNet \cite{overlapnet} and OverlapTransformer \cite{overlaptransformer} are trained using explicit overlap information on range images \cite{range_image}. OREOS \cite{oreos} proposes two separate branches: one for loop closure and other for loop detection. Unlike \cite{overlapnet, overlaptransformer, oreos} that only estimate the relative yaw between the input point clouds, we estimate the full 6-\dof relative pose.

\vspace{-2mm}
\section{Methodology}
% \vspace{-4mm}
Our goal is to develop a 6-\dof viewpoint invariant place recognition framework for 3D point clouds for \ldc. %We assume that each scene has a dominant ground plane. 
The overview of our method is shown in \cref{overall-flow}. We first canonicalize the point cloud, and then discretize it to get a \dem representation (\cref{subsec:dem_generate}). We use an \textit{autoencoder} style encoder-decoder network to learn the compressed latent representation for the \dem (\cref{subsec:dem_encoder_decoder}). The latent representation is transferred between the agents for loop detection and closure, which reduces the data bandwidth requirement. 

To achieve yaw invariance for loop detection in the compressed space, we have designed a \emph{Differentiable Yaw Transformer} (\dyt), it transforms the latent query embedding to rotationally align with the latent embedding of the database sample  (\cref{subsec:dyc}).The output of the \dyt is used for loop detection (\cref{subsec:loop_detection}). Once a loop is detected, the decoders decompress the latent \dem representation and use the decoded \dem from the query and dataset to estimate a 6-\dof relative pose for the loop closure (\cref{subsec:loop_closure}).

%the \dem representation in  the network architecture that performs place recognition using these DEMs is described in Section \ref{sec:DEM_place_rec}. In Section \ref{sec:Rel_pose_est}, the process of estimating the relative pose between two input point clouds is detailed.

\subsection{DEM Generation} 
\label{subsec:dem_generate}
\vspace{-1mm}
\dems are digital representations of an input point cloud, capturing the elevation of the terrain or overlaying objects. \dems have rich representation power, preserving the feature rich regions like edges and corners, and at the same time conserve bandwidth by allowing for aggressive compression and recovery at high quality. Moreover, unlike range images that preserve only yaw \cite{overlapnet}, \dems preserve both yaw and planar translation, making them a useful representation for 6-\dof point cloud registration.
% \vspace{-7pt}
%
% Let $\mathbf{P}_c$ be the current point cloud, whose ground place is denoted by $\mathbf{r}_c$. Let $\mathbf{P}_w$ be the world point-cloud, whose ground-plane is donated by $\mathbf{r}_w$. The world ground-plane has the normal $\mathbf{n}_w = [0,0,1]$.
%
%The \dem generation process involves two steps:
%\begin{enumerate*}[label=\textbf{(\arabic*)}]
%	\item The input point cloud could have an arbitrary 6-\dof pose. We leverage upon the dominant ground plane assumption to perform Roll and Pitch (\textit{RP}) canonicalization.
%	\item Post roll-pitch compensation, we generate the DEM by discretizing the top view into uniform cells.  
%\end{enumerate*}
% For a given input current point cloud $\mathbf{P}_c$, we estimate the relative roll ($\alpha$) and pitch ($\beta$) to align the ground planes $\mathbf{r}_c$ and $\mathbf{r}_w$. Post the roll-pitch compensation, we generate the DEM $\mathbf{D}_c$ and $\mathbf{D}_w$ of the input point cloud $\mathbf{P}_c$ by discretizing into uniform cells. 
% firstly, for a given point cloud, we estimate the roll and pitch to align the ground plane ($\mathbf{r}_c$) of the current point cloud ($\mathbf{P}_c$) to the world ground plane ($\mathbf{r}_w$). In the second step, we complete the DEM generation process by discretizing the top view of the point cloud into uniform cells in 2D.  Below we explain each of these steps in detail:

%\begin{figure}[t]
%	\includegraphics[width=\linewidth]{./Images/PlaneParameterization2.png}
%	\captionof{figure}{\footnotesize{ \textbf{Plane Parameterization}:   }\label{plane_parameteization}
%	}
%\end{figure}

\mypara{Plane Parameterization} 
Consider an input point cloud $\mathbf{P}_{c}$ with its corresponding ground plane $\mathbf{r}_c$,  and the world ground-plane $\mathbf{r_w}$. We aim to align the planes $\mathbf{r}_c$ and $\mathbf{r}_w$ by estimating the relative roll and pitch (RP) between them.  
We center the input point cloud ($\mathbf{P}_c$) and extract the ground plane $\mathbf{r}_c$ using RANSAC. The ground plane is parameterized by ($\mathbf{n}_c$, $\mathbf{C}_c$), where $\mathbf{n}_c \in \mathbf{R}^{3}$ is a unit vector perpendicular to the plane and $\mathbf{C}_c = \left\{ c^c_i \mid i = \{1...n\} \right\} \in \mathbf{R}^{n \times 3} $ is the set of points $c^c_i \in \mathbf{R}^{3}$, s.t. $c^c_i$ lies on $\mathbf{r_c}$ and $\| c^c_i \| = 1$.
The world ground plane $\mathbf{r}_w$ is parameterized similarly as ($\mathbf{n}_w$, $\mathbf{C}_w$),  where $n_w =[0,0,1]$ and $\mathbf{C}_w = \left\{ c^w_i \mid i=\{1...n\} \right\} \in \mathbf{R}^{n \times 3} $ is the set of points $c^w_i \in \mathbf{R}^{3}$, s.t. $c^w_i$ lies on $\mathbf{r_w}$ and $\| c^w_i \| = 1$.  Note that the world ground plane is not estimated through data, instead is a \textit{constructed canonical plane of reference}.
The canonicalization for roll ($\alpha$), and pitch ($\beta$) involves two steps. First we obtain a coarse estimate of ($\alpha$) and ($\beta$) by aligning the normals $\mathbf{n}_c$ and $\mathbf{n}_w$. Post that, we do a finer estimate through Iterative Closest Point (\icp).

\mypara{Coarse RP Canonicalization}
Given the normals  $\mathbf{n}_c = [n_{c,x}, n_{c,y}, n_{c,z}]$ from \ransac, and $\mathbf{n}_w = [0,0,1]$, we estimate the relative roll $\alpha$ and pitch $\beta$ by solving:
{\small
\begin{gather*}
	\begin{bmatrix} 
		0 \\0\\1\end{bmatrix} = \begin{bmatrix}
		\cos(\alpha) & 0 & \sin(\alpha) \\
		\sin(\beta)\sin(\alpha) & \cos(\beta) & -\cos(\alpha)\sin(\beta) \\
		-\cos(\beta)\sin(\alpha) & \sin(\beta) & \cos(\alpha)\cos(\beta) \\
	\end{bmatrix} 
	\begin{bmatrix}  
		n_{c,x} \\n_{c,y}\\n_{c,z}
	\end{bmatrix}.
\end{gather*}
}%
The obtained closed-form solution is given as: 
{\small
\begin{align*} 
	\alpha & = \arctan \left( \frac{-n_{c,x}}{n_{c,z}} \right)\text{,}\beta& =\arctan \left( \frac{n_{c,y}}{ n_{c,z} \cos(\alpha) - n_{c,x} \sin(\alpha) } \right)
\end{align*}
}%

% \begin{figure*}[t]
% 	\centering
% 	\begin{subfigure}[b]{0.32\linewidth}
% 		\centering
% 		\includegraphics[width=\linewidth]{./Images/Diff_yaw_can.png}
% 		\subcaption{ }
% 		\label{Diff_Yaw_can}
% 	\end{subfigure}
% 	% \hfill
% 	\begin{subfigure}[b]{0.32\linewidth}
% 		\centering
% 		\includegraphics[width=\linewidth]{./Images/LatentEmbd.png}
% 		\subcaption{   }
% 		\label{cart_to_pol_to_flip}
% 	\end{subfigure}
% 	% \hfill
% 	\begin{subfigure}[b]{0.32\linewidth}
% 		\centering
% 		\includegraphics[width=\linewidth]{./Images/Yaw_can_result.png}
% 		\subcaption{   }
% 		\label{can_demo}
% 	\end{subfigure}
% 	\caption{Differential Yaw Canonicalizer} 
% \end{figure*}

\begin{figure}
%\vspace{+2mm}
\centering
\includegraphics[width=\linewidth,trim={0 0.75cm 0 0},clip]{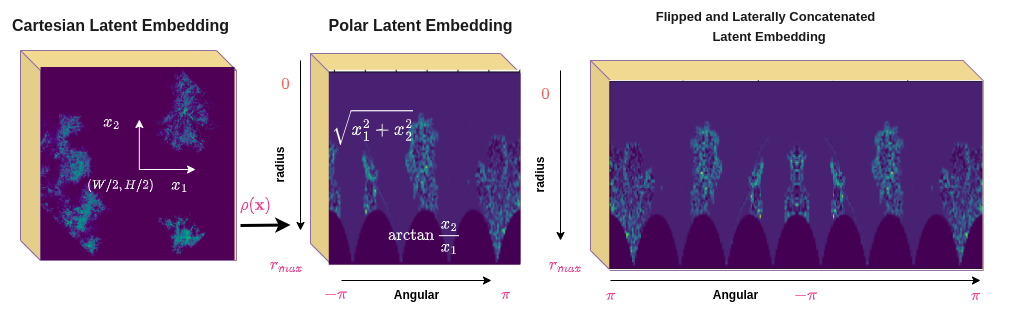}
%trim={<left> <lower> <right> <upper>}
	\caption{\footnotesize The image to the extreme left shows a sample \textit{DEM} latent space in Cartesian form. The image in the center depicts the same embedding in a polar form; the image to the right is the result of flipping and concatenation operation. }
	\label{cart_to_pol_to_flip}
\vspace{-4mm}
\end{figure}

\begin{figure}
%\vspace{+2mm}
\centering
\includegraphics[width=\linewidth]{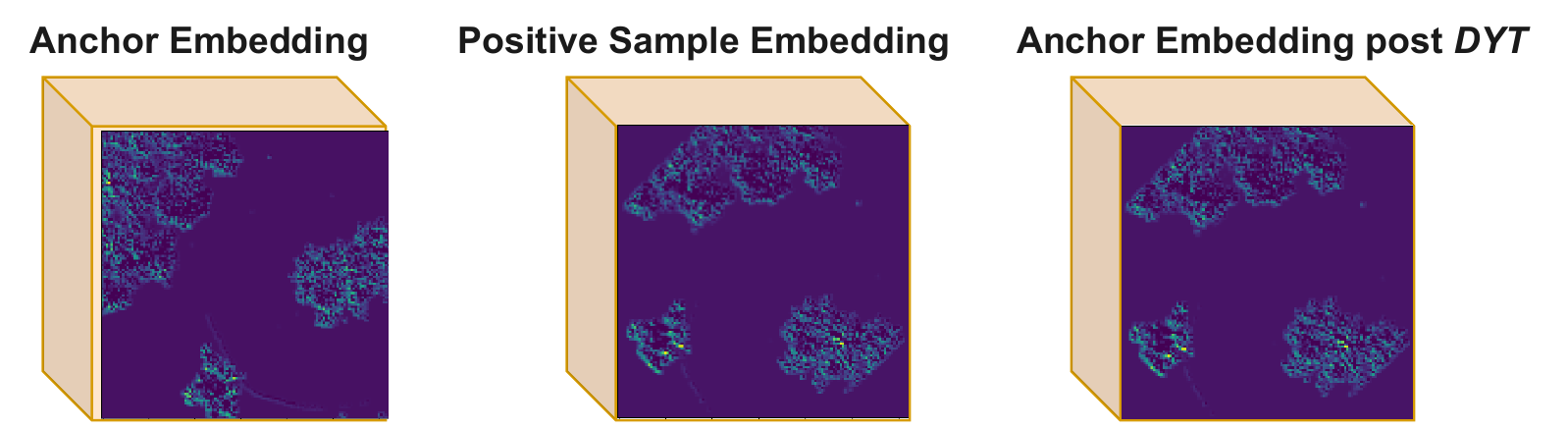}
	\caption{\footnotesize Visualization of the yaw alignment using \dyt. Note that the anchor and the positive sample are not yaw aligned initially, however post the \textit{DYT} operation the two embeddings are aligned. We show the first channel of the feature volume as binary image for ease of visualization.}
		\label{dyt_demo}
\vspace{-4mm}
\end{figure}

\mypara{Fine Grained Canonicalization with ICP} 
We use the coarse estimates of $\alpha$ and $\beta$ as described above, and refine them using \icp (initialized with coarse estimates) as:
% \[{\small
% 	R(\alpha, \beta)=  \argmin_{(\alpha,\beta)} || \mathbf{C_{w}} - R( \alpha , \beta) \mathbf{C_{c}} ||^{2} 
% }\]
\[{\tiny
	R(\alpha, \beta)=  \argmin_{(\alpha,\beta)} || \mathbf{C_{w}} - R( \alpha , \beta) \mathbf{C_{c}} ||^{2} 
}\]
\vspace{-2mm}
\mypara{Top-view Discretization} \label{sec:Top_view_discrete}
After performing the roll and pitch canonicalization, the top-view of the point cloud is discretized into uniform $2D$ grid cells to obtain the \dem $\mathbf{D}_{c}$ of point cloud $\mathbf{P}_{c}$. We define a grid $G$ of dimension $G_w \times G_h$, and resolution $d_g$. 
%The discretization process first scales the top view of the point cloud by a constant factor $s_g$ such that it fits into a fixed 2D grid with uniform resolution. 
Each grid cell $g_i \in G$ is assigned a set of points $P^g_i$ based on the resolution $d_g$, and given a height value $h^g_i = \max (h(p) \mid p \in P^g_i)$, where $h(p)$ is the height of the point $p$, thus converting a point cloud to a \dem. Such a grid representation can be readily assimilated by Deep Networks ideally suited to exploit such structural information. 
\vspace{-7pt}

\subsection{Learning Compressed Latent Representation} 
\label{subsec:dem_encoder_decoder}
We intend to use the \dems to perform the \ldc task, and an autoencoder architecture is used to generate compressed latent embeddings of a \dem. The \textit{Encoder} consists of sequentially stacked CNN layers for feature extraction, and the encoding process acts as a compressor, resulting in a feature volume (latent embedding) requiring significantly less memory to store and transmit, in comparison to the original \dem (or its corresponding point cloud). We use $\phi$ to denote the embedding, and $\phi \in R^{4 \times 125 \times 125}$. Loop detection module is designed to perform view-invariant place recognition using $\phi$. However, for loop closure, we decode $\phi$ back to the \dem before performing pose estimation. Detailed description of our encoder-decoder architecture is given in the $^{\dag}$\href{http://gurkiratsingh.me/FinderNet/}{project page}. 
\vspace{-7pt}
%It consists of three main modules;  (i) \textit{Differentiable Yaw Canonicalizer}  (\ref{Diff_can}) aligns the two DEMs such that their relative yaw angle is 0, (ii) \textit{Difference Layer} (\ref{Diff}) estimates the distance between the two DEMs which is further used by the triplet loss function, (iii) \textit{DEM Decoder} this module reconstructs the original DEM at the receiver side which is further used for point cloud registration.  Here we describe each of the modules in detail. 
%
%As described in Section  \ref{DEM_generate}, the DEM representation preserves the yaw and translation information; this is critical for applications such as  point cloud registration. Therefore, we create a DEM decoder to recover the original DEM at the receiver side. It consists of the Deconvolution layers that upsample the latent embedding and reconstructs the DEM in its original size. 
%
%For the DEM reconstruction, we train using the \textit{Mean Squared Error} (MSE) loss between the reconstructed and input DEM. The loss term is given in Eq. \ref{MSE}, where $D^x(i,j)$ refers to the $(i,j)$ pixel of either the anchor ($x=a$), positive ($x=p$) or negative ($x=n$) DEM. 
%
%\begin{equation} \label{MSE}
%	L_{MSE} = \frac{1}{HW} \sum_{x \in a,p,n} \sum_{i=0}^{H} \sum_{j=0}^{W}  || D^x(i,j) - D^x_{reconst}(i,j) || ^2
%\end{equation}

\subsection{Differentiable Yaw Transformer \textit{(DYT)} for Parameterized Grid Sampling } 
\label{subsec:dyc}
Viewpoint invariance is an important property for robust place recognition/loop detection. Previous  methods \cite{LCDNet,oreos} try to achieve this through data augmentation, where they rotate an input point cloud through randomly selected rotation angles. However, such methods do not generalize to complex sequences or large changes in viewpoint. \cite{LCDNet} acknowledges that data augmentation by itself need not be sufficient for viewpoint invariance. Therefore, we propose a \textit{Differentiable Yaw Transformer} (\dyt) module that achieves viewpoint invariance without the need for explicit data augmentation. It achieves this by receiving the latent embedding of anchor and positive (or negative) \dem (denoted as $\phi_A$, $\phi_p$, and $\phi_n$ respectively) as its input and returning the relative yaw denoted by $\psi \in R$ at the output. Then, it rotates the anchor \dem so that the relative yaw between the anchor and positive is zero. The operations within the \dyt are detailed below.  

\mypara{Cartesian to Polar Conversion (CPC)} \label{subsec:cyc}
Let $\mathbf{G}$ be a group of rotation transformations (in \sotwo) parameterized by $\psi$ s.t. $T_{\psi}: R^{d} \to R^{d}, \forall T_{\psi} \in SO(2)$. The canonical coordinate for $\mathbf {G} $ are defined such that a rotation by $T_{\psi}$ in the Cartesian coordinates appears as a translation by $\psi$ in the canonical coordinates. The polar coordinate system forms such canonical coordinates for the group of rotation transformations \cite{ET, PT}, and can be obtained from Cartesian coordinates $\mathbf{x}: (x_1, x_2)$ as:  
%
% \vspace{-6pt}
 { %\small
\begin{equation}
	\label{map1}
	\rho( \mathbf{x}) = \left(  \arctan{ \frac{x_2}{x_1} } , \sqrt{x_{1}^{2} + x_{2}^{2} } \right).
\end{equation} 
}
% \vspace{-3pt}
%
%The Cartesian to polar form conversion transforms a given anchor latent embedding  $\mathbf{\phi}_{A}$ (anchor) and $\mathbf{\phi}_{P/N}$ (positive or negative), to polar coordinates through $\rho(\phi_{A} )$ and $\rho(\phi_{P/N} )$ respectively, the conversion from cartesian to polar form of a sample latent embedding is shown in Fig. ~\ref{cart_to_pol_to_flip} (observe the images in the extreme left and in center). The input cartesian and the resulting polar space are both of dimension $4 \times 125 \times 125$  (format: number of channels $\times$ Height $\times$ Width ) .  
%
% We perform the \cpc independently for each channel of latent embedding tensor $\phi$, and generate an output tensor of the same size. 

\noindent \cpc is performed for each channel of the embedding tensor $\phi$, that results in an output tensor of the same size. The \textit{CPC} process is shown in the first two columns of Fig. ~\ref{cart_to_pol_to_flip} 

% The visualization of the \textit{CPC} process is shown in the first two columns of Fig. ~\ref{cart_to_pol_to_flip} 

\mypara{Horizontally Padding Polar Embedding} \label{subsec:horizontal_filp_and_pad}
As described above two embeddings related by a yaw rotation in the Cartesian coordinates are related by a translation after conversion to polar coordinates. However, if we try to estimate translation directly, the estimation process can only correlate between the overlapping regions. We observe that the horizontal axis of the polar latent embedding lies within the range $[ -\pi, \pi]$ and is cyclic. The cyclic property allows us to pad the embedding by copying the embedding, flipping it (the flipped embedding will be within the range $[ \pi, -\pi]$ ), and then use the flipped version to horizontally pad the embedding.  The resulting embedding is shown in \cref{cart_to_pol_to_flip} (extreme right). The operation doubles the size of the latent embedding to $4 \times 125 \times 250$, and allows us to use full embedding for translation estimation.

\mypara{Correlation Layer} \label{correlation}
After padding the polar embedding from the positive (negative) embedding, we try to locate anchor embedding in it using correlation. We implement the layer as a convolutional layer with polar latent embedding of the anchor as a kernel, and perform cross-correlation over the horizontally padded polar feature volume of the positive/negative sample.  This results in a 1D output of size $1 \times 1 \times 126$. The output of the correlation layer divides the 360 degrees of rotation into 126 bins, each of resolution 2.85 degrees (approximately). We apply softmax over the correlation score output to convert the score vector to the probability vector for various candidate translations. The predicted translation is multiplied by 2.85 to convert to predicted rotation angle. Modules similar to correlation layer have been previously used in \cite{overlapnet}, however, their setting required explicit yaw supervision, one of our contribution is to relax this requirement by formulating a \textit{soft} yaw estimation as a part of the self-supervised \dyt. This results in improved performance for large view-point changes as demonstrated by the \textit{Kitti-08} and \textit{LUF} sequences in Section \ref{loop_detect_results}. 

%The resulting rotation angle $\psi$ can be estimated using \ref{yaw_est} where Eq. \ref{softmax}  is the softmax function and $S^{i}$ refers to the $i^{th}$ element in the 126D feature vector. 
%	
%	
%	\begin{subequations}\label{yaw_est}
%		\begin{equation}
%			\psi = \sum_{ i=0 }^{126} (2.85i) a_i
%		\end{equation}
%		\begin{equation}\label{softmax}
%			a_i = \frac{S^i}{\sum_{j=0}^{126} S^j }
%		\end{equation}
%	\end{subequations}

\mypara{Rotation Sampler}  
We construct a rotation matrix $R_{\psi} \in$ \sotwo from the predicted yaw angle ($\psi$) as determined from the previous step. Similar to \cite{ST},  we use  $R_{\psi}$  to differentiably sample from the input feature volume and produce a warped output feature map, denoted as $\widehat{\phi}$. The operation is denoted as $\otimes$ in \cref{overall-flow}. Note that the operation is performed on $4 \times 125 \times 125$ dimensional embedding tensor in Cartesian coordinates. \cref{dyt_demo} depicts the result of the \dyt module, it may be seen that the anchor and positive sample do not share the same orientation at the input of \dyt. However, post-\dyt, they have same orientation. For simplicity of illustration, we only show the first channel of the $4 \times 125 \times 125$ tensor. The warped anchor tensor is sent to the next module for loop detection.

%\begin{figure*}
%\includegraphics[width=\textwidth]{./Images/Alignment_pipeline.png}
%\captionof{figure}{\footnotesize{ \textbf{Qualitative Result of Query and Recall}:   }\label{Query_Recall_res
		%}
	%}
%\end{figure*}

\vspace{-2mm}
\subsection{Loop Detection} 
\label{subsec:loop_detection}
Our pipeline achieves rotation invariance using the RP canonicalizer and DYT. Additionally, the difference layer within the loop detection module provides translation invariance and measures the similarity between the two yaw-aligned \dems.
A fully convolutional network (\cnn) is translation-equivariant. Our loop detection module consists of shared \cnn layers to extract features $F_{a} \in R^{H \times W \times C}$ from the anchor and $F_{p/n} \in R^{H \times W \times C}$ features from the positive or negative sample \dem. The difference layer takes the two feature volumes as input and computes all pairs absolute differences between the pixels. To implement all pairs absolute difference we first construct a tile tensor $T_{a} \in R^{HW \times HW \times C}$ by first reshaping $F_a$ to a $HW \times 1 \times C$ tensor, and then repeating first column in each channel by $HW$ times. Mathematically:  $ \forall i \in \{0,1,2,.., H-1\}$ and $j \in \{0,1,2,.., W-1\}$. 
\[
T_a( iW+j, k, c ) = F_a(i,j,c), \quad \forall k \in [0,HW-1].
\]
We compute $T_p$ and $T_n$ similarly, but additionally transpose each channel of the tensor at the end. This is equivalent to:
\[
T_{p/n}(k, iW+j, c) = F_{p/n}(i,j,c), \quad \forall k \in [0,HW-1].
\]
$T_a$ and $T_{p/n}$ allow us to compute all pair difference as: $F_\text{diff} = |T_{a} - T_{p/n}|$. 

%
%and $T_{p/n} \in R^{HW \times HW \times C}$ using \cref{tile1} and \cref{tile2}.   Where $i \in \{0,1,2,.., H-1\}$ and $j \in \{0,1,2,.., W-1\}$ and $k \in \{0,1,2, ..., WH -1 \}$.  Eq ~\ref{res_diff} gives the result of the difference layer and $|.|$ is a function that returns the absolute value of the element-wise difference. 
%\begin{subequations} 
%	\begin{equation}\label{tile1}
	%		T_{a}( iW+j, k,c ) = F_{a}(i,j,c)
	%	\end{equation}
%	\begin{equation} \label{tile2}
	%		T_{p/n}( k, iW+j , c) = F_{p/n}(i,j,c)
	%	\end{equation}
%	\begin{equation} \label{res_diff}
	%		F_{diff} = | T_{a} - T_{p/n}| 
	%	\end{equation}
%\end{subequations}
%
The difference layer results in a feature volume $F_\text{diff}$ that quantifies the shared information between the two \dems. $F_\text{diff}$ is passed through proposed \cnn architecture (details in the $^{\dag}$\href{http://gurkiratsingh.me/FinderNet/}{project page}), resulting in a single scalar value indicating the distance between the two \dems. A low value indicates loop detection. We train the proposed loop detection module using triplet based contrastive loss: 
\begin{equation} \label{triplet}
	\mathcal{L}_\text{triplet} = \max \left( 0, d \left(\widehat{\phi}_a, \phi_p\right) - d \left(\widehat{\phi}_a, \phi_n \right) + \xi \right),
\end{equation} \label{loss}
where $\phi$ is the \dem encoding, $\widehat{\phi}$ is the yaw aligned \dem encoding, $d$ is the distance between the two encoding computed by the loop detection module, and $\xi$ is the margin for the triplet loss. Note that during back-propagation, the \dem encoder receives gradient both from the MSE loss of the autoencoder, as well as from the above triplet loss. Whereas the decoder is trained only using the MSE loss. 
\vspace{-7pt}
%\mypara{Loss Function}
%%
%The network is expected to satisfy a twofold objective of place recognition and DEM reconstruction. For place recognition, we train the model using the triplet loss function given an \textit{Anchor} ($D^a$), \textit{Positive} ($D^p$) and \textit{Negative} ($D^n$) \dem; the triplet loss enforces the model to estimate the distance between the positive and anchor descriptors to be lower than the distance between the negative and anchor descriptors by a constant margin. The triplet loss is given by Eq. \ref{triplet} where $d(.)$ is the distance function and $f(.)$ represents the feature volume of the input DEM and  $k$ is the margin. 
%
%\begin{equation} \label{triplet}
%	L_{triplet} = \max( 0, d( f( D^a ) , f(D^p) ) - d( f( D^a ) , f(D^n) ) + k)
%\end{equation}
%
%Eq. ~\ref{loss} gives the final loss function where $w_1$ and $w_2$ are the weights that trade off the triplet and MSE loss. 
%
%\begin{equation} \label{loss}
%	L = w_1 L_{triplet} + w_2 L_{MSE}  
%\end{equation}

\subsection{Loop Closure}
\label{subsec:loop_closure}

\vspace{-2mm}
After performing the loop detection process, we estimate the \sethree rigid body transform to align the query and retrieved point cloud. This process is known as loop closure (or point cloud registration). Let $\mathbf{P}_q$ be the query point cloud, with pose $\mathbf{T}^w_q$, centered at $\mathbf{O}_q$. Let $\mathbf{P}_r$ be the retrieved point cloud, centered at origin $\mathbf{O}_r$ with a pose $\mathbf{T}^w_r$. Refer to the loop closure block in the extreme right of \cref{overall-flow}. Both  $\mathbf{T}^w_q$ and  $\mathbf{T}^w_r$(denoted in dotted blue) are in the world frame of reference, and are unknown. We aim to find the relative transformation $\mathbf{T}^q_r$ (in solid yellow) that aligns $\mathbf{P}_q$ and $\mathbf{P}_r$. To estimate the relative \sethree pose, we first estimate the relative \setwo transform between $\mathbf{P}_q$ and $\mathbf{P}_r$. Post that, its combined with the initially estimated roll and pitch canonicalization to obtain the \sethree pose estimation. 

To estimate the relative \setwo transform we decode the query and retrieved \dems from their respective encoding. Then, key-points and correspondences between the query and retrieved \dems is obtained using \cite{superpoint, superglue}. The \setwo pose is obtained from the following optimization problem
\vspace{-1mm}
\begin{equation}\label{eq:SE2_optim}
	\argmin_{\psi, \mathbf{t}^{cq}_{cr}} \| \left( \mathbf{R}(\psi)^{cq}_{cr} a_i + \mathbf{t}^{cq}_{cr} \right)  - b_i \|^{2}.
\end{equation}

Here, $a_i$, and $b_i$ are the corresponding points on the query and target \dem respectively. The rotation matrix $\mathbf{R}(\psi)^{cq}_{cr} \in$ \sotwo is parameterized by the yaw angle $\psi$ and the translation vector is denoted by $\mathbf{t}^{cq}_{cr} \in R^2$. The translation vector is scaled to the metric scale using the grid resolution $d_g$ (c.f. Top-view Discretization within \cref{subsec:dem_generate}). The yaw angle $\psi$ for the optimization is initialized using the yaw estimates from the \dyt module. Let $\mathbf{R}(\alpha_q,\beta_q)^q_{cq}$ and $\mathbf{R}(\alpha_r,\beta_r)^r_{cr}$ (shown in pink in Loop Closure module of \cref{overall-flow}) be the rotation matrices that align the query ($\mathbf{P}_q$) and retrieved point cloud ($\mathbf{P}_r$) to their respective roll-pitch compensated frames $\mathbf{O}_q$ and $\mathbf{O}_r$. To estimate the \sothree rotation matrix, we combine $\mathbf{R}^q_{cq}, \mathbf{R}^{cq}_{cr}$ and $\mathbf{R}^q_{cr}$: $\mathbf{R}^q_r = \mathbf{R}^q_{cq}\mathbf{R}^{cq}_{cr} {(\mathbf{R}^r_{cr})}^{-1}$.

We obtain the translation $\mathbf{t}^q_r$ by combining $\mathbf{t}^{cq}_{cr}$ and $d_r - d_q$: where $d_r$ and $d_q$ are the distance of the LiDAR from the estimated ground plane obtained (it is esimated along with the ground plane parameters through RANSAC).  $\mathbf{t}^q_r = [ \mathbf{t}^{cq}_{cr}(0), \mathbf{t}^{cq}_{cr}(1), d_r -d_q$.

\vspace{-2mm}
\section{Datasets and Implementation Details}

We use PyTorch, and train on a single NVIDIA GeForce GTX 1080 GPU, using a batch size of 12 and ADAM~\cite{adam} as optimizer for $200$ epochs for $8$ hours. The learning rate is initialized to $4 \times 10^{-4}$ and halved every $50$ epochs. A $50m \times 50m$ point cloud is converted to a linearly scaled \dem representation of size $500 \times 500$ pixels. The triplet margin, $k$, in \cref{triplet} is set to 0.75. Unlike \cite{LCDNet, overlapnet, soe_net, pointnetvlad, pcan }, we do not perform any augmentation on the input point clouds.  

To demonstrate the ability of the \dem to expose the underlying spatial structure and show the generalization of our method across point clouds with varying densities we choose three publicly available \lidar datasets \cite{ALITA, oxford_robocar,kitti}.  To further test the robustness of the method to 6-DOF motions we generate a synthetic dataset from a quadrotor. 

\begin{figure}[t]
\includegraphics[width=\linewidth]{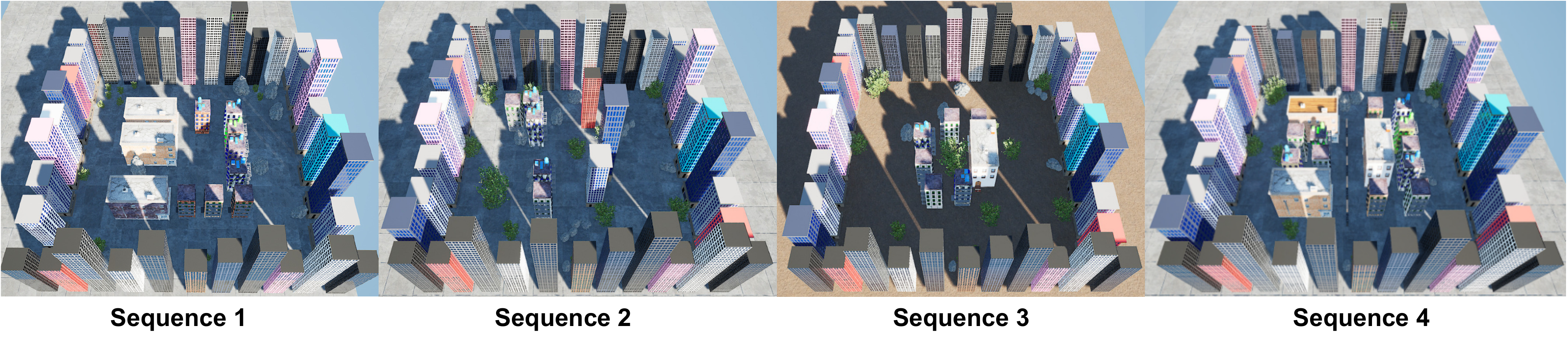}
\captionof{figure}{\footnotesize{ A glimpse of the Lidar-UrbanFly Dataset (LUF) Environment that we created. Train Data: Sequence (1,2, 3) and  Test Data Sequence 4 }}
\label{fig:unreal_6DOF}
\vspace{-8mm}
\end{figure}

\begin{enumerate*}[label=\textbf{(\arabic*)}]
    \item \textbf{KITTI} \cite{kitti}: It consists of 11 sequences, similar to \cite{LCDNet} we train on $05,06,07,09$ and test on $00$ and $08$. 
    \item \textbf{Oxford RobotCar}\cite{oxford_robocar}: The dataset consists of a total of $44$ sequences, we use train and test split of the data as recommended by \cite{pointnetvlad}. 
    \item \textbf{Lidar UrbanFly Dataset \textit{(LUF)} }: Using the \textit{Unreal} Editor \cite{unrealengine} we create a custom environment  consisting of a buildings, trees and uneven roads to evaluate \ldc methods. The environment is scanned by a $64$ channel \lidar mounted on a quadrotor in 6-DOF motion. We create four such environments as shown in Fig. ~\ref{fig:unreal_6DOF}, Sequence $(1,2,3)$ are used for training and $4$ for testing. 
    % three of which are used for training and one for testing.    
    \item \textbf{GPR}\cite{ALITA}: This dataset consists a total of $15$, we use sequence $1,2,3,4,5,6,8,9,11,12$ for training and report evaluation results on sequence $10$ and $15$.  
\end{enumerate*}
Similar to \cite{LCDNet} we consider two point clouds to form a loop when the ground truth distance between them is less than $4m$. This rule allows us to sample triplets for training, an \textit{anchor} and \textit{positive} pair is formed when the distance between their poses is less than $4m$, \textit{anchor} and \textit{negative} pair is formed when distance is between $4m$ to $10m$. All results related to Oxford Robot Car can be found on our $^{\dag}$\href{http://gurkiratsingh.me/FinderNet/}{project page}. 
\vspace{-7pt}

% We report results for KITTI \cite{kitti} and GPR \cite{ALITA} datasets in the next section. Results on Oxford Robotcar \cite{oxford_robocar} are given in the supplementary.

% The proposed network is developed using PyTorch framework and tranined on a single Nvidia GeForce GTX 1080 Ti GPU with 12GB of memory. A $50m \times 50m$ point cloud is converted to a linearly scaled \dem representation of size $500 \times 500$ pixels. The model is trained with a batch size of 12, we set $w_1$ and $w_2$ in Eq. ~\ref{loss} to be $1$ and $0.1$ respectively, ADAM optimizer with a learning rate of $4 \times 10^{-4}$ the learning rate is halved once every $50$ epochs. We train it for $200$ epochs which approximately takes about 8 hours. 

%  Similar to \cite{LCDNet} we consider two point clouds to form a loop when the ground truth distance between them is less than $4m$. This rule allows us to sample triplets for training, an \textit{anchor} and \textit{positive} pair is formed when the distance between their poses is less than $4m$, \textit{anchor} and \textit{negative} pair is formed when two scans with distance between $4m$ to $15m$ is chosen. 

% The model is tranined and tested on three publicly available detests 

\section{Experiments and Results}

% This section demonstrates the multifunctional abilities of our pipeline. We quantitatively and qualitatively demonstrate the superior efficacy of our method to perform 6-\dof \ldc in a compressed point cloud space. The experiments are aimed at testing the algorithmic robustness and include challenging scenarios such as \textit{Loop Detection} with a 6-\dof change in viewpoint and \textit{Loop Closure} without any initial guess. 
% We measure the bandwidth gained through our \dem encoding procedure. Furthermore, to measure the efficacy of our pipeline, we integrate it with \textit{LIO-SAM} \cite{lio_Sam}. Additionally we conduct a detailed abalation study to test the efficacy of individual components. Please refer to the project page for further information. 

Our pipeline has multiple functions, which we demonstrate quantitatively and qualitatively in this section. Specifically, we show that our method is highly effective in performing 6-\dof \ldc in a compressed point cloud space. We tested our algorithm's robustness through challenging scenarios such as Loop Detection with a 6-\dof change in viewpoint and Loop Closure without any initial guess. Additionally, we measure the bandwidth gained through our \dem encoding procedure and integrate our pipeline with \textit{LIO-SAM} \cite{lio_Sam} to measure its efficacy. Finally, we conducted a detailed ablation study to test the efficacy of individual components. For more information and demonstration please visit our $^{\dag}$\href{https://gurkiratsingh.me/FinderNet/}{project page}.

\vspace{-7pt}
\subsection{Loop Detection Results} \label{loop_detect_results}

% \begin{figure*}
% \includegraphics[width=\textwidth]{./Images/QualitativeResults.pdf}
% \captionof{figure}{\footnotesize{The image depicts the recall of our method on various sequences. For each of the four sequences, the point cloud in the orange box (top left) is the query point cloud, and the one within the green box (top right) is the top retrieved one. The point clouds in the red box (second row) are the second and third retrieved point clouds (left to right). This figure demonstrates the ability of our network to learn spatial priors. Note that the top-retrieved results are correct in all cases.}
% \label{Query_Recall_res}
% }
% \vspace{-4mm}
% \end{figure*}

\begin{figure}
\includegraphics[width=\linewidth]{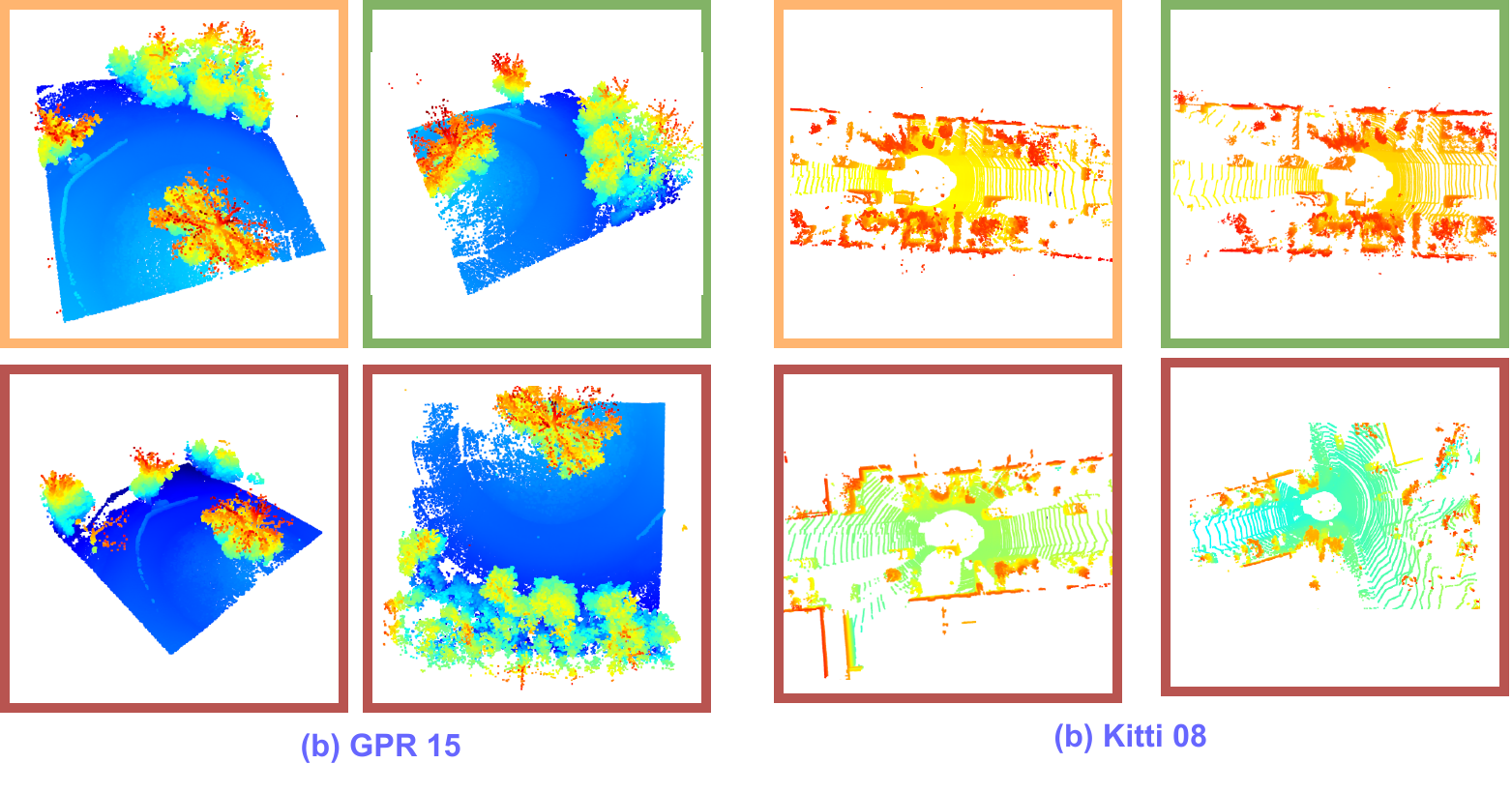}
\captionof{figure}{\footnotesize{The image depicts the recall of our method on various sequences. For each of the four sequences, the point cloud in the orange box (top left) is the query point cloud, and the one within the green box (top right) is the top retrieved one. The point clouds in the red box (second row) are the second and third retrieved point clouds (left to right). This figure demonstrates the ability of our network to learn spatial priors. Note that the top-retrieved results are correct in all cases.}
\label{Query_Recall_res}
}
\vspace{-8mm}
\end{figure}

 % We benchmark our method against multiple deep learning based methods such as \textit{LCDNet} \cite{LCDNet}, \textit{PointNetVLAD} \cite{pointnetvlad}, \textit{PCAN} \cite{pcan}, \textit{OverlapNet} \cite{overlapnet}, \textit{SOE-Net} \cite{soe_net}. For these methods we use the offical code and pre-trained model released by the authors. For fairness in comparison we retrain the model on datasets for which the pretrained model was not available.  Furthermore we benchmark our methods against a variety of methods based on handcrafted features such as ScanContext \cite{scan-context},  LiDAR-Iris \cite{lidar-iris}. 

We benchmark against \textit{LCDNet} \cite{LCDNet}, \textit{PointNetVLAD} \cite{pointnetvlad}, \textit{PCAN} \cite{pcan} and \textit{OverlapNet} \cite{overlapnet} which are the \sota deep learning based methods for loop detection, however they lack robustness to large view-point changes. Moreover, they are trained with significant augmentation which limits generalization. We use the official code and pre-trained models released by the respective authors. For fairness in comparison we retrain the model on datasets for which the pretrained model was not available i.e the retrained models are \cite{LCDNet, overlapnet, pointnetvlad,pcan, soe_net} for \textit{GPR}, \cite{pointnetvlad,soe_net, pcan} for \textit{Kitti}, \cite{LCDNet} for \textit{Oxford Robot Car}, and we retrain all models for the \textit{LUF} environment. Additionally, we benchmark against ScanContext \cite{scan-context}, a handcrafted feature based \ldc method.  

We employ \textit{Precision-Recall (PR) } curves, which is an effective metric for evaluating loop detection \cite{LCDNet, overlapnet}. To measure the \textit{Average Precision (AP)}, we follow \textit{protocol 2} suggested by \cite{LCDNet}, which is proven to be an effective benchmarking metric. Here is a brief overview of the procedure: given a query point cloud $\mathit{A}$, we compare it to all the point clouds in the database ${\mathit{B}}$. For each pair of scans $(\mathit{A}, \mathit{B_i})$, we calculate their distance (lower distance implies greater similarity) using the method described in \cref{subsec:loop_detection}. If the distance is less than a fixed threshold, it is considered a loop. We then check the Euclidean distance between the ground truth poses of the \lidar scans; if the poses are less than $4m$ apart, they are considered a \textit{true positive}, while if the distance is greater than $4m$, they are considered a \textit{false positive}. By varying this fixed threshold, we obtain multiple values of \textit{precision-recall}, which we use to plot the curve. 

% This evaluation method has been proven effective in benchmarking algorithms for challenging scenarios, such as a large change in viewpoint \cite{LCDNet}.

\begin{figure}
\includegraphics[width=\linewidth]{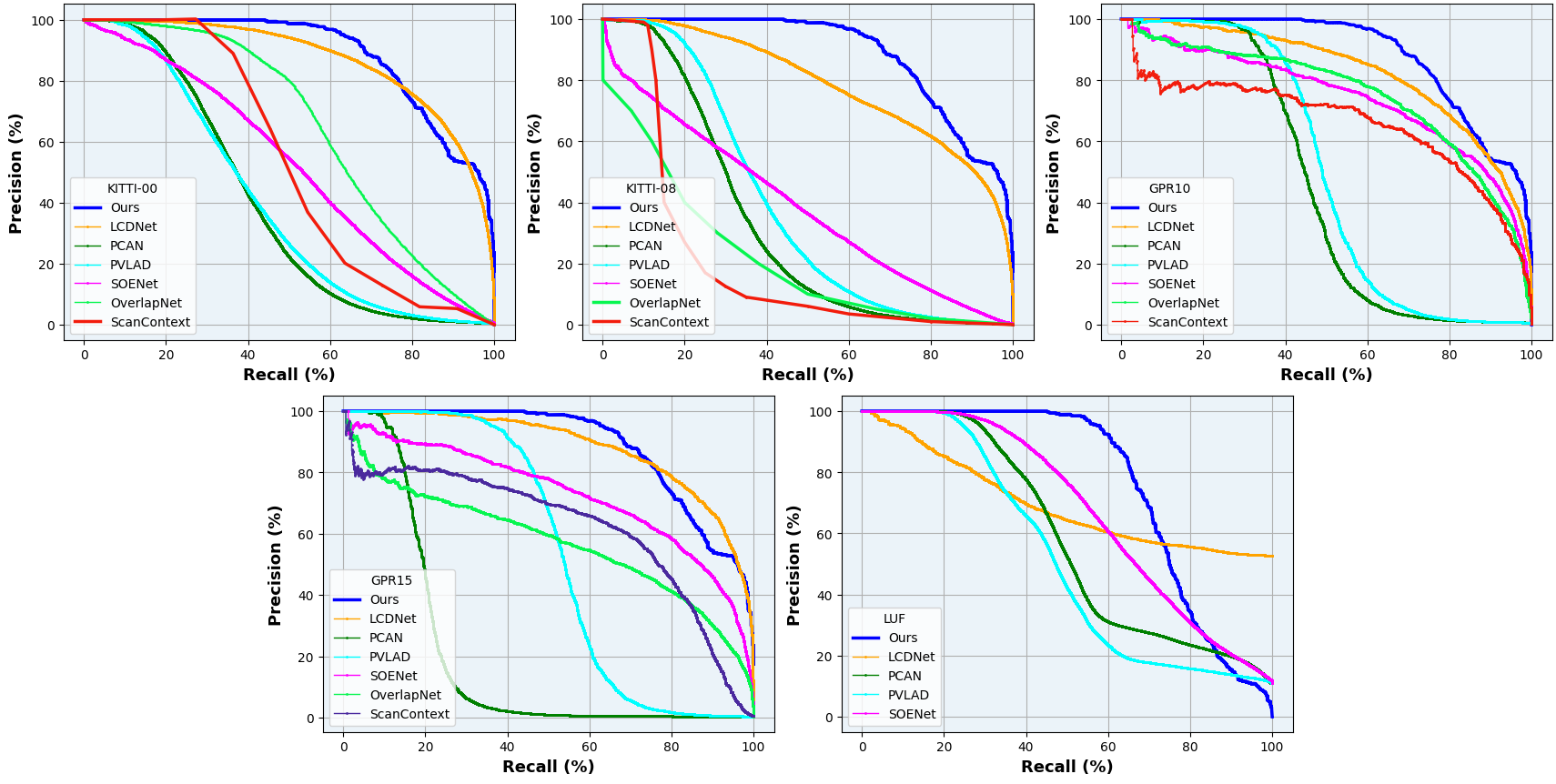}
\captionof{figure}{\footnotesize{Comparison of various loop detection algorithms on the KITTI and GPR datasets and our custom \texttt{LUF} datasets. }
\label{fig:PR_plot}
}
\vspace{-4mm}
\end{figure}

The results are presented in \cref{AP_table} and the corresponding \textit{PR} curves are depicted in \cref{fig:PR_plot}. \textit{LCDNet} \cite{LCDNet} is the \sota \ldc method on the \textit{KITTI} dataset. We observe that on \textit{KITTI08}, a challenging sequence which involves opposite views, our performance is better than \textit{LCDNet} by approximately $\mathbf{10}\%$. On the \textit{KITTI00} sequence we are second best to \sota (lower by approximately $\mathbf{1}\%$). Similarly on both the \textit{GPR} sequences $(10,15)$ our method has the highest \textit{AP} beating the closest \textit{LCDNet} by approximately $11\%$ and $4.5\%$ respectively. On the \textit{LUF} dataset, which consists of 6-DOF viewpoint changes our method outperforms  \textit{LCDNet} by $10\%$.
% Our method outperforms all the other competing benchmarks by a significant margin on both the datasets. 

In \cref{Query_Recall_res}, we present the top $3$ point clouds recalled by the \dem for a specific query to demonstrate its capability to learn the underlying spatial structure of point clouds.This demonstrates the ability of network to identify similar spatial structures. For example, In the \textit{GPR15} sequence, the presence of common structural elements like trees in both the input and recalled point clouds suggests that the underlying geometry of point clouds has been exposed by the \dem.

% To demonstrate the ability of the \dem to learn the underlying spatial structure of the point clouds, in \cref{Query_Recall_res} we show the top $3$ recalled point clouds for a given query. In addition to recovering the correct point cloud, it is important to note that the model has learnt to identify similar spatial structure. For instance, in the \textit{GPR15} sequence there are trees in both the input (query) and the matched/recalled point cloud. This shows that \dem is exposing the spatial structure of the point clouds which can be further encoded by a \cnn model.    
\vspace{-6pt}
%\vspace{-2.1mm}
\subsection{Loop Closure Evaluation} \label{loop_closure}
%\vspace{-2.1mm}

% Leveraging the benefits of the canonicalization procedure and our ability to reconstruct \dem, our pipeline allows for 6DOF registration of point clouds (as explained in section \ref{subsec:loop_closure}). In this section we compare against three types of algorithms 

In this section we compare the proposed point cloud registration method against three categories of algorithms: 
\begin{enumerate*}[label=\textbf{(\arabic*)}]
    \item \textbf{Loop Detection and only yaw estmation} approaches such as ScanContext \cite{scan-context}, OverlapNet \cite{overlapnet} only estimate the yaw and not the complete 6-\dof pose. 
    \item \textbf{Loop Detection and 6-\dof Pose Estimation:} LCDNet \cite{LCDNet} is a \sota method in the 6-\dof \ldc task and we choose to compare against it. 
    \item \textbf{Only 6-\dof relative pose estimation}: We compare with the \sota  point cloud registration technique (they do not perform loop detection), TEASER++ \cite{teaser}. We also benchmark against classical method, ICP \cite{icp}.  
\end{enumerate*}

The official code open-sourced by the authors is used to benchmark \cite{scan-context, overlapnet , teaser , LCDNet} and we implement ICP using  \textit{Open3d} \cite{open3d}. Our experimental results are shown in \cref{register_pcd}, we evaluate our method based on \textit{Average translation error (ATE)} in meters and \textit{Average Rotation Error (ARE)} in degrees. Our method has the lowest \textit{ATE} on the \textit{KITTI} dataset (both the sequences), \textit{GPR10} sequence and the \textit{LUF} dataset. It also has the lowest rotation error on \textit{KITTI08, GPR10, LUF} and \textit{GPR15}. On \textit{KITTI00} sequence, the error of \textit{FinderNet}  is higher than \textit{LCDNet} by $0.91^{0}$. Notably, methods like \cite{icp, teaser, LCDNet} process entire point clouds and will not be suitable multi agent settings due to the transfer of large point clouds required to close the loop. To improve our results, we may use outlier-resilient robust ICP formulation

\begin{table}[t]
%\small
\resizebox{\linewidth}{!}{
\begin{tabular}{||l|c|c||c|c||c||}
\hline
    \multirow{3}{*}{Method} & \multicolumn{2}{c||}{KITTI} & \multicolumn{2}{c||}{GPR} &   \multicolumn{1}{c||}{\textit{LUF}}  \\
    \cline{2-6}    
    & \makecell{KITTI-$00$} & \makecell{KITTI-$08$} & \makecell{GPR-$10$ } & \makecell{GPR-$15$ } & \makecell{Seq-4} \\
    \hline
    LCDNet~\cite{LCDNet}&$\mathbf{0.89}$ &$0.76$ &$0.82$ &$0.88$  & 0.69   \\
    OverlapNet~\cite{overlapnet}&$0.61$ &$0.22$ &$0.75$ &$0.57$  & NA  \\
    PointNetVLAD~\cite{pointnetvlad}&0.40 &0.39 &$0.50$ &$0.54$  & $0.67$ \\
    % DISCO~\cite{DISCO}&ENTER &ENTER &ENTER &ENTER &ENTER \\
    PCAN~\cite{pcan}&0.46 &0.20 & $0.39$ &$0.20$ & $0.58$  \\
    ScanContext~\cite{scan-context}&$0.49$ &$0.20$ &$0.66$ &$0.62$ & NA  \\
    % LiDAR Iris~\cite{lidar-iris}&$0.42$ &$0.17$ &ENTER &ENTER  \\
    SOE-Net~\cite{soe_net}&0.52 &0.47 &$0.74$ &$0.73$  & $0.60$  \\
    \hline
    \textit{Ours}&$0.88$ &$\mathbf{0.84}$ &$\mathbf{0.91}$ &$\mathbf{0.92}$ &   $\mathbf{0.76}$  \\
    \hline
\end{tabular}
}

\captionof{table}{ AP Comparison for loop detection. NA: Not applicable as \cite{overlapnet} and \cite{scan-context} are only for \setwo motions.} \label{AP_table}
\vspace{-7mm}
\end{table}

% \begin{table*}[ht]
% \caption{Point Cloud Registration Comparison with other Benchmark Algorithms }
% \centering  %://www.overleaf.com/project/634852ef56bdc14211498b35
% \begin{tabular}{c|c|c|c|c|c}
% \hline
%     \multirow{3}{*}{Method} & \multicolumn{2}{c}{KITTI} & \multicolumn{2}{c}{GPR} & \multicolumn{1}{c}{Oxford Robot Car}\\
%     \cline{2-6}
    
%      & \makecell{\\ KITTI-$00$} & \makecell{ \\ KITTI -$08$} & \makecell{ \\ GPR-$10$ } & \makecell{\\ GPR-$15$ } & \makecell{\\ Oxford Robot Car-Se } \\
%     \hline
%     LCDNet~\cite{LCDNet}&$0.77/1.07$ &$1.62/3.13$ &$1.44/1.14$ &$2.51$ &$1.91$ \\
%     OverlapNet*~\cite{overlapnet}&$-/3.89$ &$-/65.45$ &ENTER &ENTER &ENTER \\
%     Teaser++~\cite{teaser}&$2.72/15.85$ &$3.83/29.19$ &ENTER &ENTER &ENTER \\
%     ScanContext*~\cite{scan-context}&$-/1.92$ &$-/3.11$ &ENTER &ENTER &ENTER \\
%     LiDAR Iris*~\cite{lidar-iris}&$-/1.69$ &$-/1.84$ &ENTER &ENTER &ENTER\\
%     ICP~\cite{icp}&$2.08/8.98$ &$2.43/160.46$ &ENTER &ENTER &ENTER \\
%     \hline
%     \textit{Ours}&$0.72/1.98$ &$1.35/2.96$ &$0.95/1.85$ & $0.82/1.14$&ENTER \\
%     \hline    
% \end{tabular}\label{register_pcd}
% \end{table*}
 
\begin{table}[h!]
\small
\resizebox{\columnwidth}{!}{
\begin{tabular}{||l||c|c||c|c||c||}
\hline
    \multirow{3}{*}{Method} & \multicolumn{2}{c||}{KITTI} & \multicolumn{2}{c||}{GPR} & \multicolumn{1}{c||}{LUF} \\
    \cline{2-6}
    
     & \makecell{KITTI-$00$} & \makecell{KITTI-$08$} & \makecell{GPR-$10$ } & \makecell{GPR-$15$ } & \makecell{Seq-4} \\
    \hline
    LCDNet~\cite{LCDNet}&$0.77/\mathbf{1.07}$ &$1.62/3.13$ &$1.44/ 1.14$ &$\mathbf{0.50}/4.81$ &1.82/38.86 \\
    OverlapNet*~\cite{overlapnet}&$-/3.6$ &$-/65.29$ &$-/7.85$ &$-/6.25$ & - \\
    Teaser++~\cite{teaser}&$2.93/16.13$ &$3.24/28.98$ &$ 2.68/16.87$ &$2.47/20.34 $ & 2.05/44.12 \\
    ScanContext*~\cite{scan-context}&$-/1.89$ &$-/3.20$ &$-/4.37$ &$-/4.26$ & - \\
    % LiDAR Iris*~\cite{lidar-iris}&$-/1.69$ &$-/1.84$ &ENTER &ENTER \\
    ICP~\cite{icp}&$2.23/9.12$ &$2.31/161.16$ &2.32/7.81 &2.87/8.36 & 2.15/85.24 \\
    \hline
    \textit{Ours}&$\mathbf{0.72}/1.98$ &$\mathbf{1.35/2.96}$ &$\mathbf{0.95/0.85}$ & $0.82/\mathbf{1.14}$ & $\mathbf{1.78/35.15}$ \\
    \hline    
\end{tabular}\label{register_pcd}
}
\captionof{table}{Point Cloud Registration Comparison with \sota. Result format: \textit{TE(meters)/RE(degrees)}.  \enquote{*} are algorithms that only estimate yaw, and not directly comparable with our 6-\dof method. \enquote{-} indicates not applicable as \cite{overlapnet} and \cite{scan-context} are only for \setwo motions. } 
\vspace{-5mm}
\label{register_pcd}
\end{table}

\begin{figure}
\centering
\includegraphics[width=\linewidth]{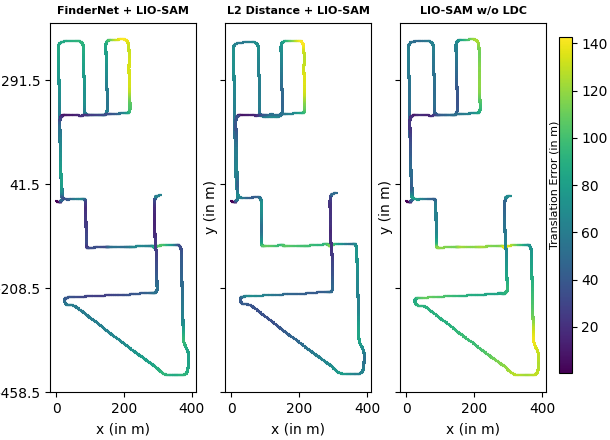}
\captionof{figure}{\footnotesize{ The figure depicts the \textit{RMSE} obtained by integrating multiple \ldc methods with \textit{LIO-SAM} \cite{lio_Sam} on kitti-08 sequences. The \textit{RMSE} of the total trajectory without \ldc (right) is $48.79m$, if the Euclidean Distance based \ldc (full \textit{LIO-SAM} center image ) an \textit{RMSE} of $35.69m$ is observed. However, integrating \textit{FinderNet} (left) the \textit{RMSE} reduces to $29.96m$. }
\label{fig:lio_sam}
}
\vspace{-7mm}
\end{figure}

\mypara{Integration with LIO-SAM}
%\vspace{-3mm}
%
We tested our pipeline's efficacy by integrating it with LIO-SAM, a state-of-the-art \lidar Inertial \slam method. LIO-SAM is a method that utilizes a factor graph based backend optimization, and has been proven to provide reliable state estimates. The results are shown in Fig. ~\ref{fig:lio_sam}, where the use of \textit{FinderNet} leads to a $16\%$ decrease in \textit{RMSE} in comparison to the \textit{LIO-SAM}'s $L2$ distance based \ldc \cite{lio_Sam}. This experiment further proves that our system can operate in real time. Refer to the $^{\dag}$\href{https://gurkiratsingh.me/FinderNet/}{project page} for the implementation details and runtime statistics.   
\vspace{-5pt}
% By integrating our pipeline with LIO-SAM, we were able to demonstrate its ability to accurately detect and close loops. Detailed implementation and results of this experiment can be found in the supplementary material.

% Similar to \cite{overlapnet, scan-context} we utilize the geometry of the factor graph for the \ldc task, for every new state $x_{i+1}$ added to the factor graph a local area of $15m$ is searched and the closest subset of possible matches is recovered. Our pipeline is used to robustly estimate a viewpoint invariant loop within from these initial set of matches. We measure the distance (as explained in section ~\ref{subsec:loop_closure}) between the query all the point clouds in the subset, if the distance is lesser than a fixed threshold it is considered as a loop and a new link would be added into the graph for optimization.  

% \begin{figure}
% \centering
% \includegraphics[width=\linewidth]{./Images/LoopDetectionVis.png}
% 	\caption{\footnotesize The Loop Detection Results on the \textit{Kitti08} sequence \tikzcircle[fill=green]{3pt}  indicates true positive \tikzcircle[fill=orange]{3pt} indicates false positives  }
% 		\label{loop_detection_kitti}
% \end{figure}
% \vspace{-1mm}
\mypara{Ablation Study}
To evaluate the performance of \textit{RP Canonicalizer} and \dyt, we conducted ablation studies. The presence of the \textit{RP-Canonicalizer} provided two critical advantages. Firstly, it allowed our method to operate on point clouds in 6-DOF motion, as demonstrated in the \ldc results on \textit{LUF} dataset (Tab. ~\ref{AP_table} and Tab. ~\ref{register_pcd}). Secondly, in the absence of the canonicalizer, we had to rely on methods such as \cite{icp, teaser} to estimate the 6-DOF relative pose. However, we observed from Tab. \ref{register_pcd} our loop closure pipeline provided improved accuracy over \cite{icp, teaser}. Furthermore, we independently study the performance of the \textit{RP-Canonicalizer}, we record that the Coarse RP Canonicalizer had an error of 3.249/4.1582 (R/P) degrees, while the fine alignment had an error of 1.2598/1.352 (R/P) degrees. 

To analyze the importance of the \dyt, we replaced it with the \textit{Spatial Transformer} \cite{ST}, which predicts the parameters for an affine transform. However, we observed that such a network led to poor performance and an \textit{AP} of $0.0551$, $0.0136$, and $0.0818$ for the \textit{Kitti-00} \cite{kitti}, \textit{LUF}, and \textit{GPR} \cite{ALITA} datasets, respectively. The \dyt was able to estimates yaw with an error of $3.12$ degrees.

Finally, we studied the bandwidth requirements and reported a compression of approximately $830$ times in comparison to the size of the original point cloud. We have provided the experimental procedure, and additional results related to the entire ablation study on our $^{\dag}$\href{https://gurkiratsingh.me/FinderNet/}{project page}.   
%\vspace{-3mm}
\vspace{-2pt}
\section{Conclusion and Future Work}
%\vspace{-2mm}
\vspace{-5pt}
We develop a novel method for 6-\dof \ldc which works on the proposed compressed pointcloud representation.  Our approach utilizes canonicalization and \dyt to achieve viewpoint invariance for large rotation angles without data-augmentation. Furthermore, unlike the previous works our method does not require data-augmentation for training. Our method demonstrates significant improvement over \sota on both real-world and simulated datasets. In future, we would like to perform resilient \ldc on dynamic scenes. 
\vspace{-10pt}

\bibliographystyle{IEEEtran}
\bibliography{references}

\end{document}